\documentclass[letterpaper]{article} 
\usepackage{aaai2027} 
\usepackage[hyphens]{url}  
\usepackage{graphicx} 
\usepackage{natbib}  
\usepackage{caption} 
\usepackage{amsmath}
\usepackage{amssymb}
\usepackage{algorithm}
\usepackage{algorithmic}
\usepackage{newfloat}
\usepackage{listings}
\DeclareCaptionStyle{ruled}{labelfont=normalfont,labelsep=colon,strut=off} 
\floatstyle{ruled}
\newfloat{listing}{tb}{lst}{}
\floatname{listing}{Listing}

\usepackage{booktabs}
\usepackage{multirow}
\usepackage{arydshln}

\title{Ground, Cover, and Refine: Evidence-Centric Frame Selection for Long-Video Question Answering}
\author{
    Fan Wei\textsuperscript{\rm 1},
    Siru Zhong\textsuperscript{\rm 5},
    Runmin Dong\textsuperscript{\rm 2}\corresponding,
    Miao Yang\textsuperscript{\rm 1},
    Zhaoyang Luo\textsuperscript{\rm 4},
    Haohuan Fu\textsuperscript{\rm 1,\rm 3,\rm 4}\corresponding
}
\affiliations{
    \textsuperscript{\rm 1}Department of Earth System Science, Tsinghua University\\
    \textsuperscript{\rm 2}School of Artificial Intelligence, Sun Yat-Sen University\\
    \textsuperscript{\rm 3}National Supercomputing Center in Shenzhen\\
    \textsuperscript{\rm 4}Tsinghua Shenzhen International Graduate School, Tsinghua University\\
    \textsuperscript{\rm 5}The Hong Kong University of Science and Technology (Guangzhou)
}

\begin{document}

\maketitle

\begin{abstract}
Long-video question answering requires identifying sparse yet critical evidence from videos containing thousands of frames under a constrained visual-token budget. Existing methods either select query-aware frames in a single pass or rely on timestamped text solely as retrieval guidance, leading to two key limitations. \textit{First}, selected frames tend to cluster around local relevance peaks, and once the budget is exhausted, omitted evidence cannot be recovered. \textit{Second}, textual and visual evidence remain weakly aligned. We propose \textbf{GCR}, a training-free framework that casts fixed-budget frame selection as a joint evidence curation problem. \textbf{\underline{G}}round converts timestamped text into temporal events, selects query-relevant real frame anchors, and renders each event text onto its temporally aligned frame. \textbf{\underline{C}}over supplements grounded events with direct visual anchors for complementary visual evidence and applies global maximal marginal relevance to preserve diverse context. \textbf{\underline{R}}efine revisits omitted temporal regions and replaces the weakest revisable context frame with a real-frame medoid---but only when the medoid offers greater evidence value. GCR maintains a fixed number of chronologically ordered frames and requires no VLM training or architectural modification. Experiments on LongVideoBench and Video-MME, across three 7B backbones and frame budgets of 8, 32, and 64, demonstrate consistent improvements in long-video QA. With the 7B LLaVA-OV backbone and 32 frames, GCR achieves 64.25\% and 62.15\% on the two benchmarks, outperforming the strongest reproduced baselines by 2.54 and 1.93 percentage points, respectively.
\end{abstract}


\begin{figure}[t]
  \centering
  \includegraphics[width=\columnwidth]{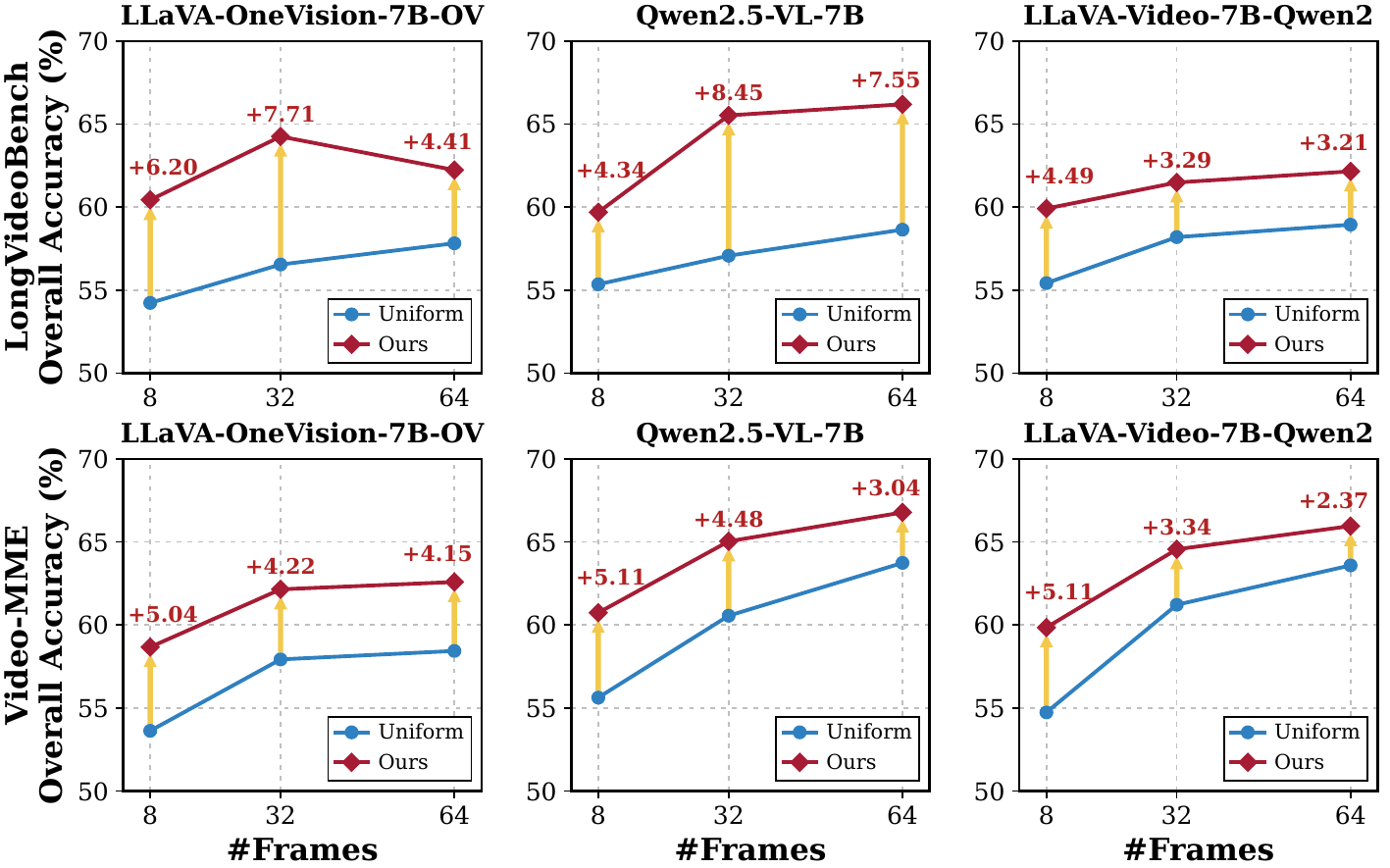}
  \caption{GCR consistently outperforms uniform sampling across VLM backbones, datasets, and frame budgets.}
  \label{fig:accuracy_comparison}
  \vspace{-1em}
\end{figure}

\begin{figure*}[t]
  \centering
  \includegraphics[width=\textwidth]{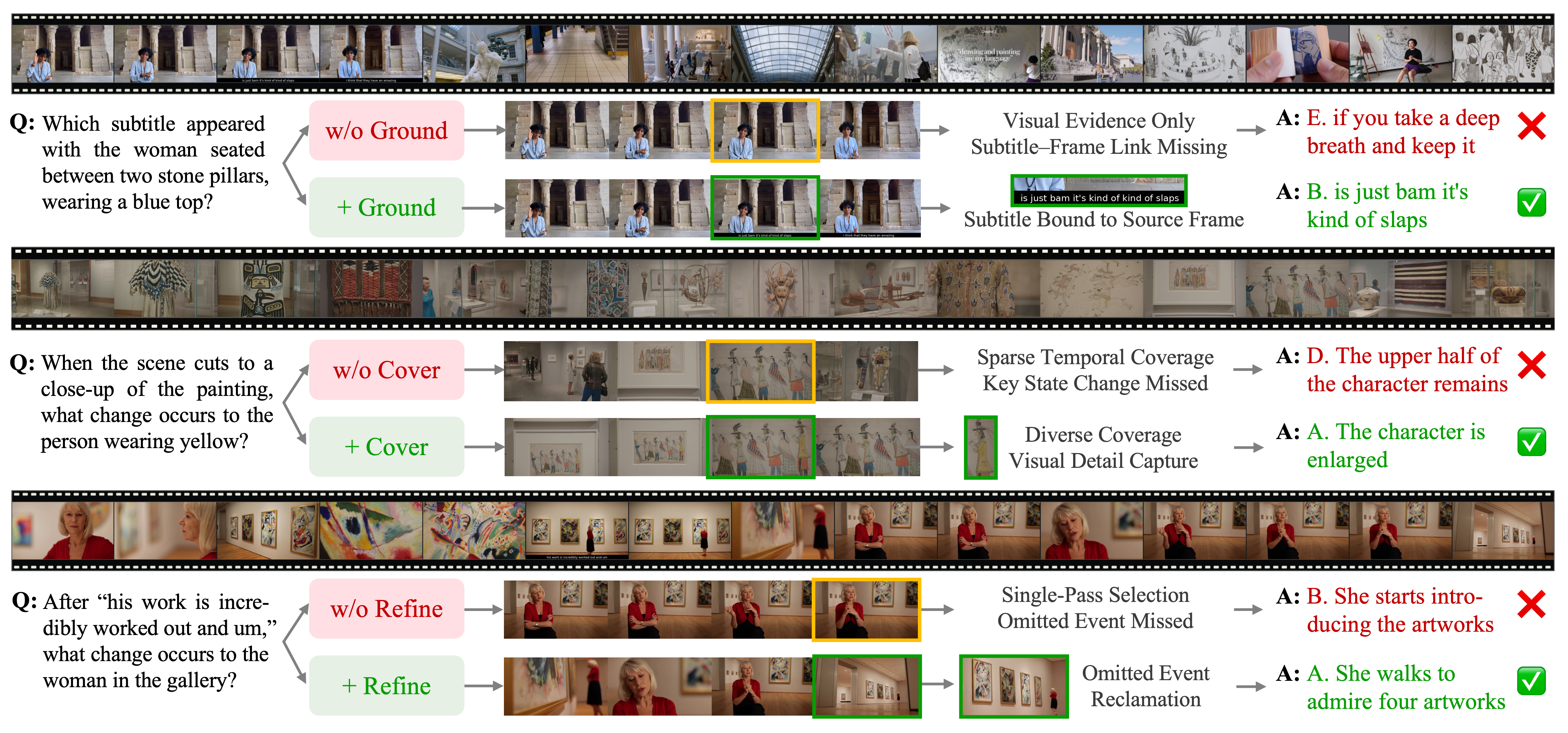}
  \caption{Qualitative examples of how \textsc{Ground}, \textsc{Cover}, and \textsc{Refine} complement each other: GCR grounds textual evidence in its corresponding visual context, captures complementary visual details, and retrieves decisive events from omitted regions.}
  \label{fig:qualitative_examples}
\end{figure*}

\begin{figure*}[t]
  \centering
  \includegraphics[width=\textwidth]{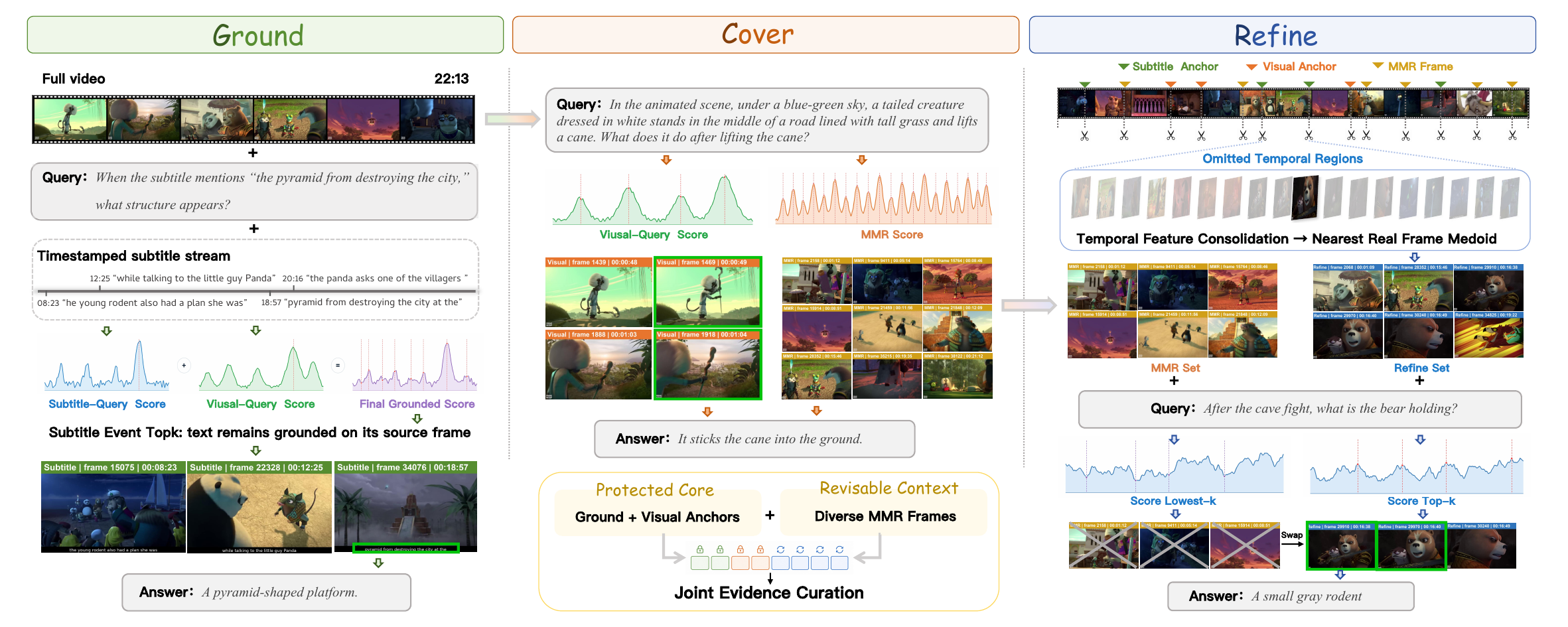}
  \caption{Overview of GCR. \textsc{Ground} binds timestamped language events to
their aligned real frame anchors, \textsc{Cover} constructs a protected
evidence core and diverse revisable context, and \textsc{Refine}
revisits omitted temporal regions to replace weak context frames with
more valuable real frame representatives while preserving the fixed
frame budget.}
  \label{fig:overiew}
\end{figure*}

\section{Introduction}

Vision-language models (VLMs) have substantially advanced
video understanding, enabling models to answer questions that require
visual perception, narrative comprehension, and temporal reasoning.
However, extending these models from short clips to long-form videos
remains fundamentally constrained by context length, GPU memory, and
inference cost. A long video may contain tens of thousands of frames,
whereas current VLMs can process only a small visual subset~\cite{LongVLM}. Sparse frame selection is
therefore an almost unavoidable preprocessing step, and the quality of the selected subset determines what evidence remains
available to the downstream VLMs. This makes long-video frame selection not merely a top-$K$
retrieval problem, but a fixed-budget joint evidence curation problem. 


Existing approaches mainly follow three routes. (1) Query-aware visual selectors improve over uniform sampling by optimizing frame-query relevance together with complementary criteria such as temporal coverage, set-level diversity, semantic boundaries, and iterative visual verification~\cite{bolt,aks,mdp3,wfssb,tstar,logic}. However, most finalize the frame set in a single pass, potentially omitting brief but informative events. (2) Language-guided multimodal selectors, such as VSI~\cite{vsi}, exploit subtitles or automatic speech recognition (ASR) for localization and score fusion, yet typically use language only as retrieval guidance or present it separately from selected frames, introducing cross-modal alignment ambiguity during downstream VLM inference. (3) Video memory and compression methods preserve long-range information through compressed tokens, latent memory, generated summaries, or explicit memory units~\cite{llamavid,moviechat,MovieChat+,MemoryCard}. These approaches introduce transformed representations or additional inputs beyond a fixed real-frame budget, often requiring architecture-specific integration or incurring extra computational cost.

Despite their differences, these three routes leave room for improvement in two aspects: more explicit grounding of textual evidence in corresponding visual observations, and greater revisability of the selected evidence set after its initial construction, as in Figure~\ref{fig:qualitative_examples}. Moreover, the set must preserve complementary visual
evidence and broad context beyond the events described by timestamped
language. This motivates the following question:
\emph{How can a fixed-budget frame set be grounded across language and vision, complementary in coverage, and revisable after its initial construction?}

To address this question, we propose \textbf{GCR}, a training-free framework that \underline{G}rounds, \underline{C}overs, and \underline{R}efines evidence for long-video question answering. GCR treats frame selection as constructing and refining a fixed-budget joint evidence set (Figure~\ref{fig:overiew}). In the \textsc{Ground} stage, subtitles or ASR transcripts serve as timestamped event indices to localize relevant visual evidence and render it onto corresponding event anchors, presenting aligned textual cues and visual observations within the same image slot. In the \textsc{Cover} stage, a dedicated visual pathway retrieves query-relevant events not covered by timestamped text. These visual anchors, together with the grounded event anchors, form a protected evidence core. Global maximal marginal relevance fills the remaining budget with diverse contextual frames, which remain revisable during subsequent refinement. The \textsc{Refine} stage revisits omitted temporal regions, represents each with a real-frame medoid, and jointly evaluates these representatives alongside contextual evidence based on query relevance, visual novelty, and local change. An omitted-region representative replaces the weakest contextual frame only when it provides greater evidence value, recovering missed evidence without increasing the VLM input budget.

As summarized in Figure~\ref{fig:accuracy_comparison}, GCR consistently outperforms uniform sampling across both benchmarks, three 7B backbones, and multiple frame budgets, demonstrating strong generalization. Ablation studies confirm that \textsc{Ground}, \textsc{Cover}, and \textsc{Refine} provide complementary gains, validating the progressive  joint evidence curation design.


Our main contributions are summarized as follows:
\begin{itemize}
    \item We formulate fixed-budget frame selection as  joint evidence curation and introduce GCR, a training-free framework that progressively grounds local events, covers complementary context, and refines the initial selection.

    \item We introduce two complementary mechanisms: event-aligned rendering, which binds timestamped language to real frames, and omitted-evidence recovery, which revisits omitted temporal regions and replaces weaker context frames when finding more valuable evidence.

    \item Extensive experiments on two long-video benchmarks, three VLM backbones, and multiple frame budgets show that our method consistently outperforms state-of-the-art approaches. Additional ablations and analyses validate its effectiveness, robustness, interpretability, and efficiency.
\end{itemize}

\section{Related Work}


\subsection{Query-Aware Frame Selection}
Query-aware frame selection improves over uniform sampling by
explicitly searching for evidence relevant to the question.
Relevance-based methods rank candidate frames using pretrained
image-text encoders, while BOLT~\cite{bolt} and AKS~\cite{aks} further balance query relevance
and temporal coverage through adaptive partitioning.
 Set-aware methods model
interactions among the selected frames. MDP$^3$~\cite{mdp3} incorporates
list-wise diversity and temporal sequentiality, whereas WFS-SB~\cite{wfssb}
detects semantic boundaries and performs diversity-aware selection
within the resulting segments. Fine-grained and agentic
approaches instead employ object-level verification, iterative
temporal search, or language-model agents to identify informative
moments~\cite{videoagent,videotree}. 
These studies demonstrate the importance of
relevance, diversity, and temporal structure. However, most
fixed-budget frame selectors finalize the selected subset after a
single construction pass, leaving omitted temporal regions
unreconsidered. GCR instead uses \textsc{Cover} to construct a
complementary initial evidence set and \textsc{Refine} to reclaim an
omitted region only when it is more valuable.

\subsection{Language-Guided Multimodal Frame Selection}
Timestamped language provides valuable temporal guidance when
answers depend on narration, dialogue, or subtitle-referred events.
VSI~\cite{vsi} combines visual search with subtitle matching, whereas Q-Gate~\cite{qgate}
routes different questions among visual and narrative scoring
streams. In these approaches,
language primarily guides frame retrieval, while the selected
images and associated text are supplied to the downstream VLM as
separate prompt components. The model must therefore recover their
fine-grained correspondence during inference. In contrast, \textsc{Ground} renders each admitted timestamped text event onto the frame it localizes,
directly presenting aligned textual and visual evidence in the same
image-based input.

\subsection{Video Memory and Compression}
Another line of work expands accessible long-video context through long-context modeling, visual-token compression, and video memory. LongVILA~\cite{longvila} extends long-context video modeling, while LLaMA-VID~\cite{llamavid} compresses each image into a compact token representation. MovieChat~\cite{moviechat} compresses dense visual tokens into sparse memory, while MovieChat+~\cite{MovieChat+} introduces question-aware memory consolidation to preserve more information from query-relevant segments. MemoryCard~\cite{MemoryCard} constructs explicit image-based memory units by rendering representative visual moments together with generated event-level video gists. These methods improve the
density or persistence of long-video evidence, but typically rely on latent memory states, generated summaries, reusable memory banks, or additional memory inputs. GCR instead keeps the downstream VLM input fixed to $B$ real frames. Through \textsc{Refine}, it reuses information from omitted regions only through traceable one-for-one replacement, without modifying the downstream architecture
or expanding its visual input.


\section{Methodology}

\subsection{Problem Formulation and Budget Allocation}
\label{sec:formulation}

We sample each video at 1 FPS to obtain a candidate frame sequence
\begin{equation}
    \mathcal{F}=\{f_t\}_{t=1}^{N},
\end{equation}
where $f_t$ denotes the $t$-th candidate frame. For a
multiple-choice question, $q$ is formed by concatenating the question
stem with all valid answer options. Let $B$ denote the maximum number
of frames that can be processed by the downstream VLM. Our goal is to
select a frame subset
\begin{equation}
    \mathcal{S}^{\star}
    =
    \arg\max_{\substack{\mathcal{S}\subseteq\mathcal{F}\\
                        |\mathcal{S}|=B}}
    U(\mathcal{S};q),
    \label{eq:ideal_objective}
\end{equation}
where $U$ denotes the downstream QA utility. Direct optimization is intractable because it requires combinatorial subset search and repeated VLM inference.

GCR allocates the budget among narrative-grounded anchors
$\mathcal{G}$, visual anchors $\mathcal{V}$, and contextual frames
$\mathcal{C}$:
\begin{equation}
\begin{gathered}
|\mathcal{G}|\leq K_g,\qquad
|\mathcal{V}|\leq K_v,\qquad
|\mathcal{C}|=B-|\mathcal{G}|-|\mathcal{V}|,\\
\mathcal{P}=\mathcal{G}\cup\mathcal{V},\qquad
\mathcal{S}_0=\mathcal{P}\cup\mathcal{C},\qquad
|\mathcal{S}_0|=B.
\end{gathered}
\label{eq:budget_allocation}
\end{equation}
The protected core $\mathcal{P}$ is never evicted, whereas at most
$K_r^{\max}$ contextual frames can be replaced during refinement.


\subsection{Ground: Narrative-Grounded Event Anchoring}
\label{sec:ground}

\paragraph{Timestamped text events.}
We construct timestamped text events from the temporal annotations
of subtitles or ASR transcripts. For each event, the associated
candidate set contains all frames whose timestamps fall within the
corresponding interval:
\begin{equation}
    \mathcal{F}_i
    =
    \{f_t\in\mathcal{F}\mid a_i\leq\tau_t\leq b_i\}.
    \label{eq:event_candidates}
\end{equation}
If no candidate falls within a very short interval, we use the frame
closest to its temporal center.

We use the frozen vision and text towers of SigLIP
~\cite{zhai2023sigmoidlosslanguageimage} as $\phi_v(\cdot)$ and $\phi_t(\cdot)$,
respectively. 
Frame-query
relevance is defined as
\begin{equation}
    s(f,q)
    =
    \cos\!\left(\phi_v(f),\phi_t(q)\right).
    \label{eq:frame_query_relevance}
\end{equation}
Each event is evaluated by textual relevance and the strongest visual
evidence within its interval:

\begin{equation}
\begin{array}{l@{\quad}c@{\quad}l}
r_i^{\mathrm{text}}
& = &
\cos\!\left(\phi_t(c_i),\phi_t(q)\right),\\
r_i^{\mathrm{vis}}
& = &
\operatorname{TopMean}_{2}
\left\{s(f,q)\mid f\in\mathcal{F}_i\right\}.
\end{array}
\label{eq:event_relevance}
\end{equation}

TopMean$_2$ reduces to Top-1 for singleton intervals.


The two event-score vectors are independently min--max normalized
within the current video. Their fusion is controlled by a
question-only visual-demand prior $d_q\in[0,1]$:
\begin{equation}
    R(e_i)
    =
    (1-d_q)\,\widetilde{r}_i^{\mathrm{text}}
    +
    d_q\,\widetilde{r}_i^{\mathrm{vis}}.
    \label{eq:event_fusion}
\end{equation}
We empirically determine $d_q$ according to the visual and textual reliance implied by keywords within each question stem. Detailed rules are provided in the supplementary material.

\paragraph{Unique event anchors.}
Events are processed in descending order of $R(e_i)$ until the
capacity $K_g$ is reached.

\begin{equation}
    f_i^{\star}
    =
    \arg\max_{\substack{f\in\mathcal{F}_i\\
                        f\notin\mathcal{G}}}
    s(f,q).
    \label{eq:event_anchor}
\end{equation}

\paragraph{Within-slot grounding.}
The original event text is rendered directly onto its selected frame:
\begin{equation}
    \widehat{f}_i
    =
    \operatorname{Render}(f_i^{\star},c_i).
    \label{eq:within_slot}
\end{equation}
Rendering binds each timestamped text to its localized frame within a fixed bottom band, without enlarging the canvas or appending text to the prompt. Only grounded anchors are rendered, and all other selected frames remain unchanged.


\subsection{Cover: Complementary Evidence Construction}
\label{sec:cover}

\paragraph{Visual evidence anchoring.}
Timestamped text provides strong temporal localization, but it does
not exhaustively describe every object, attribute, action, or state
required for question answering. Therefore we reserve a visual pathway:
\begin{equation}
    \mathcal{V}
=
\operatorname{TopK}_{K_v}
\bigl(
\mathcal{F}\setminus\mathcal{G};
s(\cdot,q)
\bigr),
\end{equation}
where $\operatorname{TopK}_{K_v}(\mathcal A;s)$ returns the
$K_v$ elements of $\mathcal A$ with the highest scores under $s$.


\paragraph{Context diversity coverage.}
High-confidence anchors may still concentrate around a few local
moments. Starting from the entire protected core
$\mathcal{S}\leftarrow\mathcal{P}$, we therefore use global MMR
~\cite{carbonell1998use} to fill the remaining budget. At each step,
we select
\begin{equation}
    f^{\star}
    =
    \arg\max_{f\in\mathcal{F}\setminus\mathcal{S}}
    \left[
        0.5 s(f,q)
        -
        0.5\max_{z\i
        n\mathcal{S}}
        \cos\!\left(\phi_v(f),\phi_v(z)\right)
    \right].
    \label{eq:mmr}
\end{equation}
The first term preserves query relevance, while the second penalizes
redundancy with respect to all selected core and context frames.
Selection continues until $|\mathcal{S}|=B$. The newly selected frames form the revisable context set
$\mathcal{C}$, which provides diverse and broad contextual support.


\subsection{Refine: Omitted Evidence Reclamation}
\label{sec:reclaim}

Refine revisits omitted temporal regions and admits their
evidence only when it is more valuable than the weakest revisable
context frame.

\paragraph{Omitted temporal regions.}

Let the timestamps in $\mathcal{S}_0$ be sorted as
\begin{equation}
    0=\tau_0<\tau_1<\cdots<\tau_B<\tau_{B+1}=T_V,
\end{equation}
where $T_V$ is the video duration. The unselected candidates between each pair of adjacent boundaries form an omitted temporal region:
\begin{equation}
    g_j
    =
    \{f_t\in\mathcal{F}\setminus\mathcal{S}_0
      \mid \tau_j<\tau_t<\tau_{j+1}\}.
    \label{eq:residual_gap}
\end{equation}

Let $\kappa(f,z)=\cos(\phi_v(f),\phi_v(z))$ denote the
visual similarity between two frames. For each non-empty gap, we measure query relevance, novelty with
respect to the initial set, and local visual change:
\begin{equation}
\begin{array}{@{}l@{\;}c@{\;}l@{}}
\operatorname{Rel}(g)
& = &
\operatorname{TopMean}_{2}
\{s(f,q)\mid f\in g\},
\\[0.15em]
\operatorname{Novel}(g)
& = &
1-\displaystyle
\max_{f\in g,\,z\in\mathcal{S}_0}
\kappa(f,z),
\\[0.15em]
\operatorname{Change}(g)
& = &
\displaystyle
\frac{1}{|g|-1}
\sum_t
\left[1-\kappa(f_t,f_{t+1})\right].
\end{array}
\label{eq:gap_descriptors}
\end{equation}
For a singleton gap, we define $\operatorname{Change}(g)=0$.


\paragraph{Real frame medoid.}
We summarize each region in feature space and retain the real frame
closest to its normalized centroid:
\begin{equation}
h_g
=
\arg\max_{f\in g}
\cos\!\left(
\phi_v(f),
\operatorname{Norm}
\left(
\frac{1}{|g|}
\sum_{z\in g}\phi_v(z)
\right)
\right).
\label{eq:residual_medoid}
\end{equation}
Thus, feature aggregation introduces no hidden memory input, and the
downstream VLM still receives a real RGB frame.


\paragraph{Unified evidence value.}
For each context frame $c\in\mathcal{C}$, we compute analogous
frame-level relevance, novelty, and change scores following
Eq.~\eqref{eq:gap_descriptors}. Real frame medoids and
context frames are jointly min--max normalized along each dimension.
Their unified value is
\begin{equation}
V(u)
=
\frac{
\widetilde{\operatorname{Rel}}(u)
+
\widetilde{\operatorname{Novel}}(u)
+
\widetilde{\operatorname{Change}}(u)
}{3}.
\label{eq:evidence_value}
\end{equation}
Equal weighting reflects their complementary roles in identifying informative evidence from omitted regions.



\paragraph{Temporal omission gate and adaptive eviction.}
Let $G_{\max}$ be the duration of the largest omitted temporal region.
Refinement is activated only if
\begin{equation}
\frac{G_{\max}}{T_V}\geq\tau_g.
\label{eq:gap_gate}
\end{equation}
where $\tau_g$ is a threshold hyperparameter.

\begin{table*}[t]
  \centering
  \caption{\textbf{Downstream task evaluation results on two benchmarks.} All accuracy scores (\%) are from our rigorously controlled replication. \textsuperscript{*} denotes uniform frame sampling, and \textsuperscript{\textdagger} denotes methods with training. \textsuperscript{$\ddagger$} denotes that the original grounding cache is unavailable. We regenerate it once using frozen Qwen2.5-VL-7B and reuse it across all matched settings. 
  \textcolor{red}{Red} upward arrows indicate the absolute gain of ours over the second-best method.}
  \label{tab:main_results}
  \renewcommand{\arraystretch}{1.15} 
  \resizebox{\textwidth}{!}{%
  \begin{tabular}{@{} l c cccc @{\hspace{\tabcolsep}\vrule width \arrayrulewidth\hspace{\tabcolsep}} cccc @{}}
    \hline
    \multirow{3}{*}{\textbf{Model and Strategy}} & \multirow{3}{*}{\textbf{Frame}}
    & & \multicolumn{2}{c}{\textbf{\textsc{LongVideoBench}}} &
    & & \multicolumn{2}{c}{\textbf{\textsc{Video-MME}}} & \\
    & & \textbf{Short} & \textbf{Medium} & \textbf{Long} & \textbf{Overall} & \textbf{Short} & \textbf{Medium} & \textbf{Long} & \textbf{Overall} \\
    & & {\scriptsize <3m} & {\scriptsize 3-15m} & {\scriptsize 15-60m} & & {\scriptsize <2min} & {\scriptsize 4-30min} & {\scriptsize >30min} & \\
    \hline

    LLaVA-OV\textsuperscript{*} & 8 & 67.59 & 52.67 & 46.81 & 54.23 & 64.78 & 51.56 & 44.56 & 53.63 \\
    LLaVA-OV + AKS (CVPR25)~\cite{aks} & 8 & 63.99 & 56.31 & 53.72 & 57.29 & 67.00 & 55.56 & 48.67 & 57.07 \\
    LLaVA-OV + VSI\textsuperscript{$\ddagger$} (CVPR26 findings)~\cite{vsi} & 8 & 65.65 & 52.67 & 47.87 & 54.15 & 62.11 & 50.56 & 45.89 & 52.85 \\
    LLaVA-OV + TSPO\textsuperscript{\textdagger} (AAAI26)~\cite{tang2025tspotemporalsamplingpolicy} & 8 & 67.04 & 57.52 & 51.24 & 57.44 & 65.43 & 56.19 & 46.78 & 56.10 \\
    LLaVA-OV + WFS-SB (CVPR26)~\cite{wfssb} & 8 & 68.98 & 59.47 & \textbf{53.90} & 59.69 & 67.11 & \textbf{59.00} & 47.22 & 57.78 \\
    \textbf{LLaVA-OV + GCR (ours)} & 8 & \textbf{72.02} {\scriptsize\scalebox{0.85}{\textcolor{red}{$\uparrow$3.04}}} & \textbf{62.14} {\scriptsize\scalebox{0.85}{\textcolor{red}{$\uparrow$2.67}}} & 51.77 & \textbf{60.43} {\scriptsize\scalebox{0.85}{\textcolor{red}{$\uparrow$0.74}}} & \textbf{67.67} {\scriptsize\scalebox{0.85}{\textcolor{red}{$\uparrow$0.56}}} & 57.33 & \textbf{51.00} {\scriptsize\scalebox{0.85}{\textcolor{red}{$\uparrow$2.33}}} & \textbf{58.67} {\scriptsize\scalebox{0.85}{\textcolor{red}{$\uparrow$0.89}}} \\
    \hline
    
    LLaVA-OV\textsuperscript{*} & 32 & 70.64 & 54.85 & 48.76 & 56.54 & 69.22 & 55.78 & 48.78 & 57.93 \\
    LLaVA-OV + AKS (CVPR25)~\cite{aks} & 32 & 70.64 & 57.77 & 57.27 & 61.03 & 69.11 & 57.44 & 51.56 & 59.37 \\
    LLaVA-OV + VSI\textsuperscript{$\ddagger$} (CVPR26 findings)~\cite{vsi} & 32 & 69.25 & 55.34 & 51.42 & 57.44 & 68.56 & 52.44 & 46.89 & 55.96 \\
    LLaVA-OV + TSPO\textsuperscript{\textdagger} (AAAI26)~\cite{tang2025tspotemporalsamplingpolicy} & 32 & 70.91 & 60.92 & 54.96 & 61.11 & 69.58 & 57.55 & 50.00 & 59.01 \\
    LLaVA-OV + WFS-SB (CVPR26)~\cite{wfssb} & 32 & 71.75 & 59.22 & 57.09 & 61.71 & 69.89 & \textbf{61.33} & 49.44 & 60.22 \\
    \textbf{LLaVA-OV + GCR (ours)} & 32 & \textbf{73.96} {\scriptsize\scalebox{0.85}{\textcolor{red}{$\uparrow$2.21}}} & \textbf{65.05} {\scriptsize\scalebox{0.85}{\textcolor{red}{$\uparrow$4.13}}} & \textbf{57.45} {\scriptsize\scalebox{0.85}{\textcolor{red}{$\uparrow$0.18}}} & \textbf{64.25} {\scriptsize\scalebox{0.85}{\textcolor{red}{$\uparrow$2.54}}} & \textbf{70.67} {\scriptsize\scalebox{0.85}{\textcolor{red}{$\uparrow$0.78}}} & 61.00 & \textbf{54.78} {\scriptsize\scalebox{0.85}{\textcolor{red}{$\uparrow$3.22}}} & \textbf{62.15} {\scriptsize\scalebox{0.85}{\textcolor{red}{$\uparrow$1.93}}} \\
    \hline
    
    Qwen2.5-VL-7B\textsuperscript{*} & 8 & 68.42 & 53.64 & 48.23 & 55.35 & 65.22 & 53.33 & 48.33 & 55.63 \\
    Qwen2.5-VL-7B + AKS (CVPR25)~\cite{aks} & 8 & 66.76 & 59.22 & \textbf{53.37} & 58.79 & 64.67 & 53.33 & 49.56 & 55.85 \\
    Qwen2.5-VL-7B + VSI\textsuperscript{$\ddagger$} (CVPR26 findings)~\cite{vsi} & 8 & 65.93 & 55.10 & 48.76 & 55.35 & 63.33 & 53.00 & 47.78 & 54.70 \\
    Qwen2.5-VL-7B + TSPO\textsuperscript{\textdagger} (AAAI26)~\cite{tang2025tspotemporalsamplingpolicy} & 8 & 67.04 & 59.71 & \textbf{53.37} & 59.01 & 64.53 & 52.48 & 49.67 & 55.54 \\
    Qwen2.5-VL-7B + WFS-SB (CVPR26)~\cite{wfssb} & 8 & \textbf{68.70} & 59.95 & 52.66 & 59.24 & 66.89 & 55.89 & 49.89 & 57.56 \\
    \textbf{Qwen2.5-VL-7B + GCR (ours)} & 8 & 67.87 & \textbf{62.14} {\scriptsize\scalebox{0.85}{\textcolor{red}{$\uparrow$2.19}}} & 52.66 & \textbf{59.69} {\scriptsize\scalebox{0.85}{\textcolor{red}{$\uparrow$0.45}}} & \textbf{68.67} {\scriptsize\scalebox{0.85}{\textcolor{red}{$\uparrow$1.78}}} & \textbf{58.67} {\scriptsize\scalebox{0.85}{\textcolor{red}{$\uparrow$2.78}}} & \textbf{54.89} {\scriptsize\scalebox{0.85}{\textcolor{red}{$\uparrow$5.00}}} & \textbf{60.74} {\scriptsize\scalebox{0.85}{\textcolor{red}{$\uparrow$3.18}}} \\
    \hline
    
    Qwen2.5-VL-7B\textsuperscript{*} & 32 & 71.19 & 56.80 & 48.23 & 57.07 & 71.33 & 59.89 & 50.44 & 60.56 \\
    Qwen2.5-VL-7B + AKS (CVPR25)~\cite{aks} & 32 & 74.52 & 61.41 & \textbf{59.75} & 64.25 & \textbf{72.67} & 60.67 & 54.44 & 62.59 \\
    Qwen2.5-VL-7B + VSI\textsuperscript{$\ddagger$} (CVPR26 findings)~\cite{vsi} & 32 & 68.70 & 60.44 & 52.48 & 59.31 & 68.22 & 58.56 & 50.11 & 58.96 \\
    Qwen2.5-VL-7B + TSPO\textsuperscript{\textdagger} (AAAI26)~\cite{tang2025tspotemporalsamplingpolicy} & 32 & 74.79 & 62.14 & 57.62 & 63.65 & 71.60 & 60.14 & 55.44 & 62.37 \\
    Qwen2.5-VL-7B + WFS-SB (CVPR26)~\cite{wfssb} & 32 & 75.07 & 63.59 & 59.57 & 65.00 & 71.00 & 62.11 & 54.00 & 62.37 \\
    \textbf{Qwen2.5-VL-7B + GCR (ours)} & 32 & \textbf{75.35} {\scriptsize\scalebox{0.85}{\textcolor{red}{$\uparrow$0.28}}} & \textbf{65.78} {\scriptsize\scalebox{0.85}{\textcolor{red}{$\uparrow$2.19}}} & 59.04 & \textbf{65.52} {\scriptsize\scalebox{0.85}{\textcolor{red}{$\uparrow$0.52}}} & 72.56 & \textbf{64.56} {\scriptsize\scalebox{0.85}{\textcolor{red}{$\uparrow$2.45}}} & \textbf{58.00} {\scriptsize\scalebox{0.85}{\textcolor{red}{$\uparrow$2.56}}} & \textbf{65.04} {\scriptsize\scalebox{0.85}{\textcolor{red}{$\uparrow$2.45}}} \\
    \hline
    LLaVA-Video-7B-Qwen2\textsuperscript{*} & 8  & 64.82 & 54.85 & 49.82 & 55.42 & 67.33 & 49.78 & 47.11 & 54.74 \\
    \textbf{LLaVA-Video-7B-Qwen2 + GCR (ours)} & 8 & \textbf{68.70} {\scriptsize\scalebox{0.85}{\textcolor{red}{$\uparrow$3.88}}} & \textbf{61.41} {\scriptsize\scalebox{0.85}{\textcolor{red}{$\uparrow$6.56}}} & \textbf{53.19} {\scriptsize\scalebox{0.85}{\textcolor{red}{$\uparrow$3.37}}} & \textbf{59.91} {\scriptsize\scalebox{0.85}{\textcolor{red}{$\uparrow$4.49}}} & \textbf{70.44} {\scriptsize\scalebox{0.85}{\textcolor{red}{$\uparrow$3.11}}} & \textbf{56.22} {\scriptsize\scalebox{0.85}{\textcolor{red}{$\uparrow$6.44}}} & \textbf{52.89} {\scriptsize\scalebox{0.85}{\textcolor{red}{$\uparrow$5.78}}} & \textbf{59.85} {\scriptsize\scalebox{0.85}{\textcolor{red}{$\uparrow$5.11}}} \\
    LLaVA-Video-7B-Qwen2\textsuperscript{*} & 32 & 70.08 & 58.01 & 50.71 & 58.19 & 74.11 & 57.44 & 52.11 & 61.22 \\
    \textbf{LLaVA-Video-7B-Qwen2 + GCR (ours)} & 32 & \textbf{71.75} {\scriptsize\scalebox{0.85}{\textcolor{red}{$\uparrow$1.67}}} & \textbf{60.44} {\scriptsize\scalebox{0.85}{\textcolor{red}{$\uparrow$2.43}}} & \textbf{55.67} {\scriptsize\scalebox{0.85}{\textcolor{red}{$\uparrow$4.96}}} & \textbf{61.48} {\scriptsize\scalebox{0.85}{\textcolor{red}{$\uparrow$3.29}}} & \textbf{74.22} {\scriptsize\scalebox{0.85}{\textcolor{red}{$\uparrow$0.11}}} & \textbf{62.22} {\scriptsize\scalebox{0.85}{\textcolor{red}{$\uparrow$4.78}}} & \textbf{57.22} {\scriptsize\scalebox{0.85}{\textcolor{red}{$\uparrow$5.11}}} & \textbf{64.56} {\scriptsize\scalebox{0.85}{\textcolor{red}{$\uparrow$3.34}}} \\
    LLaVA-Video-7B-Qwen2\textsuperscript{*} & 64 & 70.64 & 58.01 & 52.13 & 58.94 & 75.78 & 61.00 & 54.00 & 63.59 \\
    \textbf{LLaVA-Video-7B-Qwen2 + GCR (ours)} & 64 & \textbf{71.47} {\scriptsize\scalebox{0.85}{\textcolor{red}{$\uparrow$0.83}}} & \textbf{61.41} {\scriptsize\scalebox{0.85}{\textcolor{red}{$\uparrow$3.40}}} & \textbf{56.74} {\scriptsize\scalebox{0.85}{\textcolor{red}{$\uparrow$4.61}}} & \textbf{62.15} {\scriptsize\scalebox{0.85}{\textcolor{red}{$\uparrow$3.21}}} & \textbf{75.89} {\scriptsize\scalebox{0.85}{\textcolor{red}{$\uparrow$0.11}}} & \textbf{63.22} {\scriptsize\scalebox{0.85}{\textcolor{red}{$\uparrow$2.22}}} & \textbf{58.78} {\scriptsize\scalebox{0.85}{\textcolor{red}{$\uparrow$4.78}}} & \textbf{65.96} {\scriptsize\scalebox{0.85}{\textcolor{red}{$\uparrow$2.37}}} \\
    \hdashline
    Qwen2.5-VL-7B\textsuperscript{*} & 64 & 68.42 & 59.22 & 51.95 & 58.64 & \textbf{75.11} & 63.11 & 53.00 & 63.74 \\
    \textbf{Qwen2.5-VL-7B + GCR (ours)} & 64 & \textbf{78.67} {\scriptsize\scalebox{0.85}{\textcolor{red}{$\uparrow$10.25}}} & 66.75 {\scriptsize\scalebox{0.85}{\textcolor{red}{$\uparrow$7.53}}} & 57.80 {\scriptsize\scalebox{0.85}{\textcolor{red}{$\uparrow$5.85}}} & \textbf{66.19} {\scriptsize\scalebox{0.85}{\textcolor{red}{$\uparrow$7.55}}} & 74.56 & \textbf{66.44} {\scriptsize\scalebox{0.85}{\textcolor{red}{$\uparrow$3.33}}} & \textbf{59.33} {\scriptsize\scalebox{0.85}{\textcolor{red}{$\uparrow$6.33}}} & \textbf{66.78} {\scriptsize\scalebox{0.85}{\textcolor{red}{$\uparrow$3.04}}} \\
    \hdashline
    LLaVA-OV\textsuperscript{*} & 64 & 68.70 & 56.80 & 51.60 & 57.82 & 69.89 & 56.89 & 48.56 & 58.44 \\
    \textbf{LLaVA-OV + GCR (ours)} & 64 & \textbf{72.02} {\scriptsize\scalebox{0.85}{\textcolor{red}{$\uparrow$3.32}}} & \textbf{61.41} {\scriptsize\scalebox{0.85}{\textcolor{red}{$\uparrow$4.61}}} & \textbf{56.56} {\scriptsize\scalebox{0.85}{\textcolor{red}{$\uparrow$4.96}}} & \textbf{62.23} {\scriptsize\scalebox{0.85}{\textcolor{red}{$\uparrow$4.41}}} & \textbf{70.33} {\scriptsize\scalebox{0.85}{\textcolor{red}{$\uparrow$0.44}}} & \textbf{61.89} {\scriptsize\scalebox{0.85}{\textcolor{red}{$\uparrow$5.00}}} & \textbf{55.56} {\scriptsize\scalebox{0.85}{\textcolor{red}{$\uparrow$7.00}}} & \textbf{62.59} {\scriptsize\scalebox{0.85}{\textcolor{red}{$\uparrow$4.15}}} \\

    \hline

  \end{tabular}%
  }
\end{table*}

Real frame medoids are processed in descending order of $V$. At each
step, we identify the weakest context frame and compute the
replacement margin:
\begin{equation}
c^\star=\arg\min_{c\in\mathcal{C}}V(c),
\qquad
\Delta(h,c^\star)=V(h)-V(c^\star).
\label{eq:replacement_margin}
\end{equation}

A replacement is accepted only when the omitted-region representative
has positive normalized relevance, novelty, and change scores, and
$\Delta(h,c^\star)>0$. We perform at most $K_r^{\max}$ one-for-one
exchanges and stop once the best remaining candidate has a
non-positive margin. Consequently, the
protected core is preserved, the final budget remains exactly $B$,
and $\mathcal{S}^\star=\mathcal{S}_0$ when no candidate is admitted.
The final frames are sorted chronologically before VLM inference.







\section{Experiments}
We conduct extensive experiments to validate GCR, addressing four research questions:

\begin{itemize}
    \item\textbf{RQ1.} Does GCR consistently outperform representative frame selection methods across datasets, VLM backbones, video durations, and frame budgets?
    \item\textbf{RQ2.} How do individual GCR stages contribute to performance, and how should textual and visual evidence be integrated?
    \item\textbf{RQ3.} How does GCR perform across different question categories?
    \item\textbf{RQ4.} How sensitive is GCR to budget allocation, and what computational overhead does it introduce?
\end{itemize}

We describe the experimental setup below, then organize results around these four questions.

\subsection{Experimental Setup}
\label{sec:setup}
GCR is evaluated as a training-free test-time frame selector. All
downstream answering VLMs remain frozen, and no model parameters are
updated.

\paragraph{Datasets.}
We evaluate \textsc{GCR} on \textbf{LongVideoBench}~\cite{wu2024longvideobench} and \textbf{Video-MME}~\cite{fu2025video}, both of which
provide timestamped textual information required for evaluating
language-grounded visual evidence. LongVideoBench contains videos
of up to 60 minutes, while Video-MME covers multimodal perception
and reasoning across diverse video durations. For Video-MME, subtitles are not incorporated into the downstream VLM prompt.

\paragraph{Baselines.} 
To situate GCR within the current landscape, we compare it with representative methods. All approaches are evaluated under identical frame budgets, downstream VLMs and decoding configurations. 
Detailed descriptions and adaptation protocols are provided in the Supplementary Material.

\paragraph{Implementation Details.}
For downstream QA, we employ LLaVA-OneVision-Qwen2-7B-OV (LLaVA-OV)~\cite{llava-ov-Qwen2-7B}, Qwen2.5-VL-7B-Instruct~\cite{Qwen2.5-VL}, and LLaVA-Video-7B-Qwen2~\cite{zhang2410video} as downstream VLMs. For Qwen2.5-VL-7B-Instruct, we set max\_pixels $= 12{,}845{,}056$ and cap each decoded video frame at video\_frame\_max\_pixels $= 602{,}112$. We evaluate
GCR under frame budgets of $B\in\{8,32,64\}$. Across all datasets,
VLMs, and frame budgets, we define the Ground, Visual, and Refine ratios as $\alpha_g=K_g/B$, $\alpha_v=K_v/B$, and
$\alpha_r=K_r^{\max}/B$, respectively. Unless otherwise specified,
we use $(\alpha_g,\alpha_v,\alpha_r)=(1/4,1/8,1/8)$ and assign the
remaining initial budget to MMR context completion. The same
allocation is used without dataset- or backbone-specific tuning. All experiments were conducted using four NVIDIA A800 GPUs.

\subsection{Overall Performance}
\label{sec:4.1}
As shown in Table~\ref{tab:main_results}, GCR consistently achieves
the best overall performance across different VLM backbones and frame
budgets on both LongVideoBench and Video-MME. With only 32 frames,
GCR improves LLaVA-OV to 64.25\% on LongVideoBench and
62.15\% on Video-MME, surpassing the strongest reproduced baselines
by 2.54 and 1.93 points, respectively. The improvements become more
significant on challenging long-duration videos: with a 64-frame
budget, GCR gains 5.85 points on LongVideoBench-Long and 6.33 points
on Video-MME-Long over Qwen2.5-VL-7B. 


\subsection{Ablation Study}

\paragraph{Stage contributions.}
Table~\ref{tab:ablation_modalities} presents leave-one-stage-out ablations.
Removing any stage of GCR reduces overall performance under both 8- and 32-frame budgets, confirming that the three stages play complementary roles. Among them, removing
\textsc{Cover} causes the largest degradation, lowering accuracy by 2.91 and 3.29 percentage points, respectively. This shows that localized anchors must be complemented
with diverse global context. Removing \textsc{Ground} also yields consistent drops of 1.94 and 2.40 points, validating the importance of timestamped-text event localization and aligned rendering. \textsc{Refine} provides modest but positive overall gains of 0.22 and 0.45 percentage points. This is consistent with its role as a
conservative final correction: after \textsc{Ground} and
\textsc{Cover} construct a strong initial evidence set,
\textsc{Refine} revisits omitted regions to recover additional useful
evidence without increasing the frame budget.


\begin{table}[tbhp]
  \centering
  \caption{Ablation study on the individual contributions of the three modules on LongVideoBench using LLaVA-OV.}
  \label{tab:ablation_modalities}
  \renewcommand{\arraystretch}{1.15}
  \resizebox{\columnwidth}{!}{%
  \begin{tabular}{c|cccc|cccc}
    \hline
    \multirow{2}{*}{\shortstack{\textbf{Model}\\\textbf{Variant}}}
    & \multicolumn{4}{c|}{\textbf{8 Frames} (\%)}
    & \multicolumn{4}{c}{\textbf{32 Frames} (\%)} \\
    & \textbf{Short} & \textbf{Med.} & \textbf{Long} & \textbf{Overall}
    & \textbf{Short} & \textbf{Med.} & \textbf{Long} & \textbf{Overall} \\
    \hline
    Uniform baseline & 67.59 & 52.67 & 46.81 & 54.23 & 70.64 & 54.85 & 48.76 & 56.54 \\
    w/o Ground & 70.36 & 57.28 & 51.77 & 58.49 & 70.91 & 61.89 & 56.03 & 61.85 \\
    w/o Cover & 70.36 & 56.31 & 50.18 & 57.52 & 73.13 & 60.44 & 53.55 & 60.96 \\
    w/o Refine & 69.53 & 60.19 & \textbf{54.26} & 60.21 & 73.96 & 64.56 & 56.74 & 63.80 \\
    \hline
    \textbf{Full GCR} & \textbf{72.02} & \textbf{62.14} & 51.77 & \textbf{60.43} & \textbf{73.96} & \textbf{65.05} & \textbf{57.45} & \textbf{64.25} \\
    \hline
  \end{tabular}%
  }
\end{table}

\paragraph{Visual and textual evidence integration.}

To isolate how timestamped text should be integrated with visual evidence, we compare six input configurations while keeping the VLM, frame budget, selected frames, and subtitle content fixed. For the comparison between visual input without subtitles and aligned overlay, both variants use 32 frames with identical spatial resolution, visual token count, prompt token count, and theoretical Reader FLOPs. Aligned overlay renders an average of 63.22 subtitle words onto 7.10 selected
frames per QA. As shown in Table~\ref{tab:subtitle_input_formats}, appending subtitles as a global prompt or frame-tagged text yields 61.03\% and 60.73\%, both below the visual-only baseline of 61.41\%. This suggests that text-side subtitle injection places an additional burden on the model to associate textual evidence with the corresponding visual slots and may introduce cross-modal alignment ambiguity. 

In contrast, rendering each subtitle onto its corresponding frame achieves the best accuracy of 64.25\%, improving upon the visual-only baseline by 2.84 percentage points. The blank-band control performs similarly to the visual-only baseline (61.26\%), confirming that the improvement does not arise from the added image region or layout modification. Moreover, aligned subtitle overlay outperforms shuffled overlay by 1.65 percentage points, demonstrating that the gain depends not only on subtitle semantics but also on correct subtitle frame correspondence. These results validate our aligned overlay design: binding textual evidence to its visual context at the pixel level provides a more effective and reliable multimodal representation than supplying subtitles through a separate textual prompt.

\begin{table}[thpb]
  \centering
  \caption{Ablation of visual and textual evidence integration on
LongVideoBench using LLaVA-OV ($B=32$). Gains are measured relative to
the visual-only configuration in percentage points.}
  \label{tab:subtitle_input_formats}
  \renewcommand{\arraystretch}{1.1}
  \resizebox{\columnwidth}{!}{%
  \begin{tabular}{lcc}
    \toprule
    \textbf{Input Configuration} & \textbf{Accuracy (\%)} & \textbf{Gain (pp)} \\
    \midrule
    Visual-only & 61.41 & 0.00 \\
    Blank subtitle band & 61.26 & -0.15 \\
    Global subtitle prompt & 61.03 & -0.37 \\
    Frame-tagged subtitle prompt & 60.73 & -0.67 \\
    Shuffled subtitle overlay & 62.60 & +1.20 \\
    \textbf{Aligned subtitle overlay} & \textbf{64.25} & \textbf{+2.84} \\
    \bottomrule
  \end{tabular}%
  }
\end{table}

\paragraph{Qualitative examples.}
Figure~\ref{fig:qualitative_examples} 
provides representative qualitative
evidence for the effectiveness of the three GCR stages. In each case,
GCR accurately localizes and captures the decisive information missed by the
baseline, enabling the downstream VLM to produce the correct answer.
These examples complement the quantitative ablations and demonstrate
the practical value of \textsc{Ground}, \textsc{Cover}, and
\textsc{Refine}.

\subsection{Category-wise Analysis}

\begin{figure}[tbhp]
  \centering
  \includegraphics[width=\columnwidth]{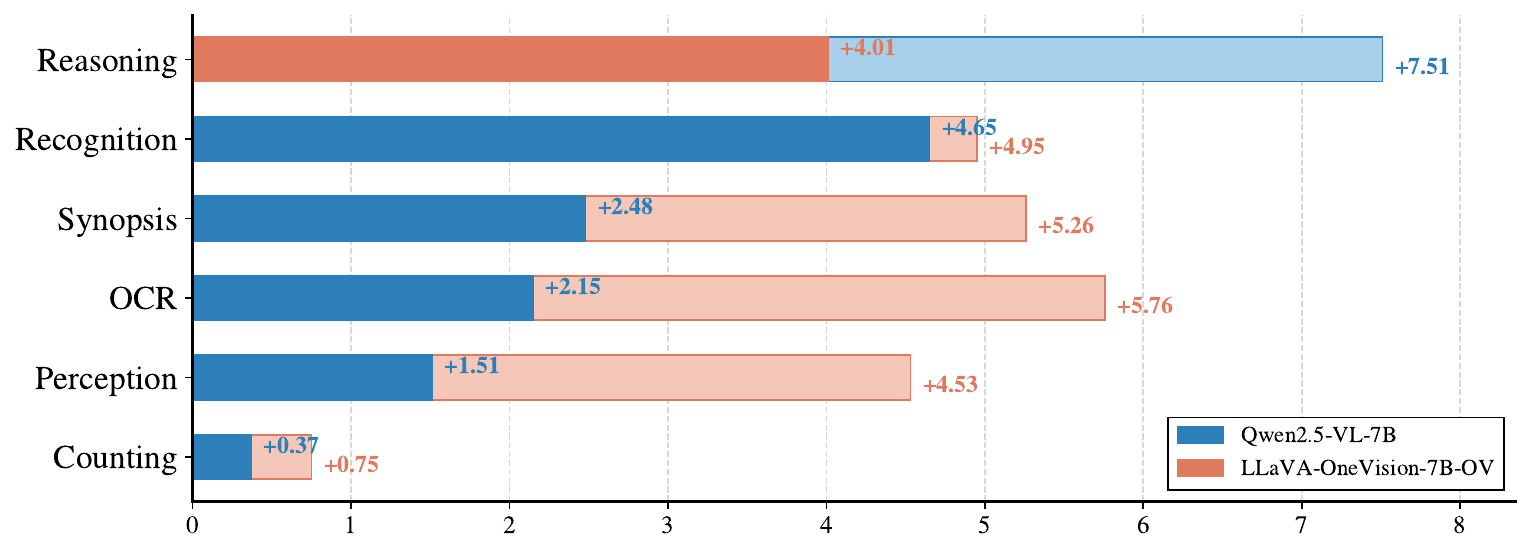}
  \caption{Category-wise accuracy gains of GCR on Video-MME using Qwen2.5-VL-7B and LLaVA-OV ($B=32$).}
  \label{fig:category_gain_samewidth}
\end{figure}

Figure~\ref{fig:category_gain_samewidth} shows that GCR is particularly effective on reasoning, recognition, OCR, and synopsis questions. For reasoning, GCR improves Qwen2.5-VL-7B and LLaVA-OV by 7.51 and 4.01 percentage points, respectively. This advantage is consistent with our aligned overlay design, which binds narrative cues to their corresponding visual
evidence, and with \textsc{Refine}, which recovers brief causal or temporal
transitions missed during initial selection. Recognition and OCR also improve substantially, as the direct visual pathway preserves query-relevant objects, actions, and text-bearing frames. For synopsis questions, MMR-based global context coverage distributes
the frame budget across complementary scenes, providing a more
representative view of the full video rather than overconcentrating on a few local relevance peaks. GCR also delivers positive gains on perception and counting across both VLMs.

\subsection{Sensitivity and Efficiency}

\paragraph{Budget sensitivity.}

\begin{figure}[tbhp]
  \centering
  \includegraphics[width=\columnwidth]{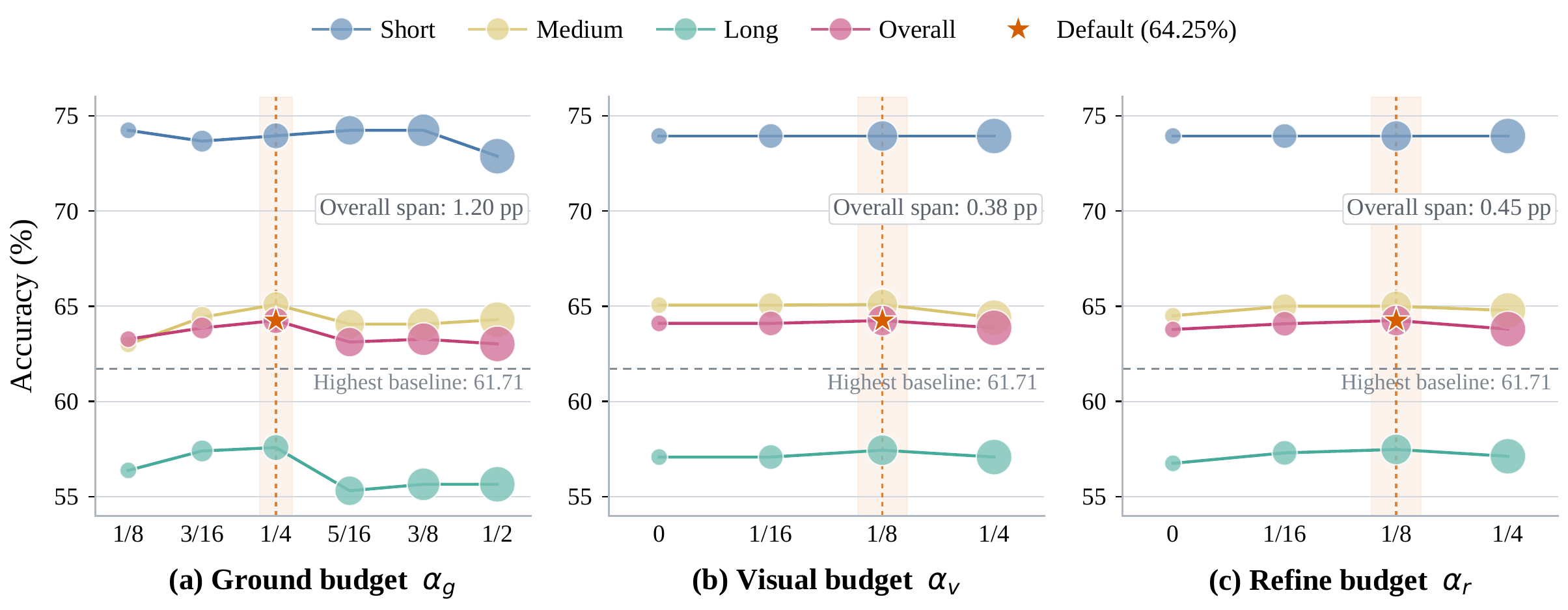}
  \caption{Sensitivity analysis of the Ground, visual coverage, and Refine budget ratios on LongVideoBench using LLaVA-OV ($B=32$).}
  \label{fig:budget_sensitivity}
\end{figure}

As shown in Figure~\ref{fig:budget_sensitivity}, GCR remains stable across all three budget allocations under a 32-frame budget. Overall
accuracy varies by only 1.20, 0.38, and 0.45 percentage points
when changing the ground, visual anchor, and maximum refine ratios, respectively, and all tested configurations remain above the
strongest baseline. The default
$(\alpha_g,\alpha_v,\alpha_r)=(1/4,1/8,1/8)$ achieves the best
overall result. More aggressive visual or refine allocations
slightly reduce accuracy, indicating that localized evidence and
omitted-region correction should complement rather than replace
diverse global context. These results show that our default
allocation is selected from a broad stable region rather than a
narrowly tuned optimum.

\paragraph{Efficiency.}
As shown in Table~\ref{tab:efficiency_comparison}, GCR achieves the highest accuracy while maintaining competitive selector-side cost.
Compared with the strongest baseline WFS-SB, GCR improves accuracy by 2.54 percentage points with a modest 1.593\,s increase in total latency, while using less peak GPU memory (2.921 vs.\ 3.180\,GiB). These results indicate that the main cost lies in full-video
decoding and feature encoding, whereas GCR's query-aware scoring and
selection require only 0.141 seconds per QA. Moreover, the Refine stage is particularly lightweight: it improves
accuracy from 63.80\% to 64.25\% while adding only 0.107\,s of selection latency on average. This confirms that revisiting omitted
regions provides an effective correction at negligible additional cost.

\begin{table}[tbph]
  \centering
  \caption{Efficiency comparison on LongVideoBench using LLaVA-OV
  ($B=32$). Latency is reported in seconds per QA. D.+Feat.\ denotes 1-FPS video decoding and visual/text feature
encoding. Score+Sel.\ denotes the complete candidate scoring and frame selection process, including Refine.}
  \label{tab:efficiency_comparison}
  \renewcommand{\arraystretch}{1.15}
  \setlength{\tabcolsep}{3.5pt}
  \resizebox{\columnwidth}{!}{%
  \begin{tabular}{c|ccc|c|c}
    \hline
    \multirow{2}{*}{\textbf{Method}}
    & \multicolumn{3}{c|}{\textbf{Latency (s/QA)}}
    & \multirow{2}{*}{\shortstack{\textbf{Peak Mem.}\\\textbf{(GiB)}}}
    & \multirow{2}{*}{\shortstack{\textbf{Acc.}\\\textbf{(\%)}}} \\
    \cline{2-4}
    & \textbf{D.+Feat.} & \textbf{Score+Sel.}
      & \textbf{Total}
      & & \\
    \hline
    AKS
      & 33.087 & 0.003 & 33.090 & 2.814 & 61.03 \\
    WFS-SB
      & 27.245 & 0.648 & 27.893 & 3.180 & 61.71 \\
    \hline
    \textbf{GCR}
      & 29.346 & 0.141 & 29.487 & 2.921
      & \textbf{64.25} \\
    \hline
  \end{tabular}%
  }
\end{table}

\section{Conclusion}
Our results show that effective long-video preprocessing requires more
than selecting the top-$K$ individual RGB frames. Under a fixed frame budget, performance depends on jointly selecting complementary visual
observations, grounding timestamped language in the corresponding source frames, and keeping the selected evidence set revisable so that
initially omitted regions can be reconsidered. GCR operationalizes this principle and delivers consistent gains across benchmarks, VLM
backbones, and frame budgets without additional training or architectural modification. Ablation, sensitivity, and efficiency
analyses further show that the three stages are complementary and that the framework is robust and lightweight.

\paragraph{Limitations.}
GCR relies on timestamped subtitles or ASR transcripts and currently has limited applicability to videos without usable timestamped text. Future work will explore grounding with audio and visual cues to improve robustness and counting performance.

\bibliography{main}

\begin{thebibliography}{24}
\providecommand{\natexlab}[1]{#1}

\bibitem[{Bai et~al.(2025)Bai, Chen, Liu, Wang, Ge, Song, Dang, Wang, Wang, Tang, Zhong, Zhu, Yang, Li, Wan, Wang, Ding, Fu, Xu, Ye, Zhang, Xie, Cheng, Zhang, Yang, Xu, and Lin}]{Qwen2.5-VL}
Bai, S.; Chen, K.; Liu, X.; Wang, J.; Ge, W.; Song, S.; Dang, K.; Wang, P.; Wang, S.; Tang, J.; Zhong, H.; Zhu, Y.; Yang, M.; Li, Z.; Wan, J.; Wang, P.; Ding, W.; Fu, Z.; Xu, Y.; Ye, J.; Zhang, X.; Xie, T.; Cheng, Z.; Zhang, H.; Yang, Z.; Xu, H.; and Lin, J. 2025.
\newblock Qwen2.5-VL Technical Report.
\newblock arXiv:2502.13923.

\bibitem[{Carbonell and Goldstein(1998)}]{carbonell1998use}
Carbonell, J.; and Goldstein, J. 1998.
\newblock The use of MMR, diversity-based reranking for reordering documents and producing summaries.
\newblock In \emph{Proceedings of the 21st annual international ACM SIGIR conference on Research and development in information retrieval}, 335--336.

\bibitem[{Chen et~al.(2026)Chen, Zeng, Luo, Xie, Lin, Ji, Zhang, and Zheng}]{wfssb}
Chen, W.; Zeng, Y.; Luo, Y.; Xie, T.; Lin, L.; Ji, J.; Zhang, Y.; and Zheng, X. 2026.
\newblock Wavelet-based Frame Selection by Detecting Semantic Boundary for Long Video Understanding.
\newblock In \emph{Proceedings of the IEEE/CVF Conference on Computer Vision and Pattern Recognition (CVPR)}, 24052--24061.

\bibitem[{Chen et~al.(2025)Chen, Xue, Li, Hu, Zhu, Li, Fang, Tang, Yang, Liu et~al.}]{longvila}
Chen, Y.; Xue, F.; Li, D.; Hu, Q.; Zhu, L.; Li, X.; Fang, Y.; Tang, H.; Yang, S.; Liu, Z.; et~al. 2025.
\newblock Longvila: Scaling long-context visual language models for long videos.
\newblock In \emph{International Conference on Learning Representations}, volume 2025, 18227--18246.

\bibitem[{Fu et~al.(2025)Fu, Dai, Luo, Li, Ren, Zhang, Wang, Zhou, Shen, Zhang, Chen, Li, Lin, Zhao, Li, Xu, Zheng, Chen, Shan, He, and Sun}]{fu2025video}
Fu, C.; Dai, Y.; Luo, Y.; Li, L.; Ren, S.; Zhang, R.; Wang, Z.; Zhou, C.; Shen, Y.; Zhang, M.; Chen, P.; Li, Y.; Lin, S.; Zhao, S.; Li, K.; Xu, T.; Zheng, X.; Chen, E.; Shan, C.; He, R.; and Sun, X. 2025.
\newblock Video-MME: The First-Ever Comprehensive Evaluation Benchmark of Multi-modal LLMs in Video Analysis.
\newblock In \emph{Proceedings of the IEEE/CVF Conference on Computer Vision and Pattern Recognition (CVPR)}, 24108--24118.

\bibitem[{Guo et~al.(2025)Guo, Chen, Wang, He, Xu, Ye, Sun, and Xiong}]{logic}
Guo, W.; Chen, Z.; Wang, S.; He, J.; Xu, Y.; Ye, J.; Sun, Y.; and Xiong, H. 2025.
\newblock Logic-in-Frames: Dynamic Keyframe Search via Visual Semantic-Logical Verification for Long Video Understanding.
\newblock arXiv:2503.13139.

\bibitem[{He et~al.(2026)He, Hong, Li, Guo, Hu, and Xiong}]{vsi}
He, J.; Hong, M.; Li, J.; Guo, W.; Hu, X.; and Xiong, H. 2026.
\newblock VSI: Visual-Subtitle Integration for Keyframe Selection to Enhance Long Video Understanding.
\newblock In \emph{Proceedings of the IEEE/CVF Conference on Computer Vision and Pattern Recognition (CVPR) Findings}, 9003--9012.

\bibitem[{Li et~al.(2024)Li, Zhang, Guo, Zhang, Li, Zhang, Zhang, Zhang, Li, Liu, and Li}]{llava-ov-Qwen2-7B}
Li, B.; Zhang, Y.; Guo, D.; Zhang, R.; Li, F.; Zhang, H.; Zhang, K.; Zhang, P.; Li, Y.; Liu, Z.; and Li, C. 2024.
\newblock LLaVA-OneVision: Easy Visual Task Transfer.
\newblock arXiv:2408.03326.

\bibitem[{Li, Wang, and Jia(2024)}]{llamavid}
Li, Y.; Wang, C.; and Jia, J. 2024.
\newblock Llama-vid: An image is worth 2 tokens in large language models.
\newblock In \emph{European Conference on Computer Vision}, 323--340. Springer.

\bibitem[{Liu et~al.(2025)Liu, Zhao, Xu, and Ghanem}]{bolt}
Liu, S.; Zhao, C.; Xu, T.; and Ghanem, B. 2025.
\newblock BOLT: Boost Large Vision-Language Model Without Training for Long-Form Video Understanding.
\newblock \emph{2025 IEEE/CVF Conference on Computer Vision and Pattern Recognition (CVPR)}, 3318--3327.

\bibitem[{Song et~al.(2024)Song, Chai, Wang, Zhang, Zhou, Wu, Chi, Guo, Ye, Zhang, Lu, Hwang, and Wang}]{moviechat}
Song, E.; Chai, W.; Wang, G.; Zhang, Y.; Zhou, H.; Wu, F.; Chi, H.; Guo, X.; Ye, T.; Zhang, Y.; Lu, Y.; Hwang, J.-N.; and Wang, G. 2024.
\newblock MovieChat: From Dense Token to Sparse Memory for Long Video Understanding.
\newblock In \emph{Proceedings of the IEEE/CVF Conference on Computer Vision and Pattern Recognition (CVPR)}, 18221--18232.

\bibitem[{Song et~al.(2026)Song, Chai, Ye, Hwang, Li, and Wang}]{MovieChat+}
Song, E.; Chai, W.; Ye, T.; Hwang, J.-N.; Li, X.; and Wang, G. 2026.
\newblock MovieChat+: Question-Aware Sparse Memory for Long Video Question Answering.
\newblock \emph{IEEE Transactions on Pattern Analysis and Machine Intelligence}, 48(1): 374--389.

\bibitem[{Sun et~al.(2025)Sun, Lu, Wang, Chen, Xu, Luo, Zhang, and Li}]{mdp3}
Sun, H.; Lu, S.; Wang, H.; Chen, Q.-G.; Xu, Z.; Luo, W.; Zhang, K.; and Li, M. 2025.
\newblock Mdp3: a Training-Free Approach for List-Wise Frame Selection in Video-Llms.
\newblock \emph{2025 IEEE/CVF International Conference on Computer Vision (ICCV)}, 24090--24101.

\bibitem[{Tang et~al.(2026)Tang, Han, Sun, Zhou, Zhang, Wei, Yuan, Zhang, Xu, and Sun}]{tang2025tspotemporalsamplingpolicy}
Tang, C.; Han, Z.; Sun, H.; Zhou, S.; Zhang, X.; Wei, X.; Yuan, Y.; Zhang, H.; Xu, J.; and Sun, H. 2026.
\newblock Tspo: Temporal sampling policy optimization for long-form video language understanding.
\newblock In \emph{Proceedings of the AAAI Conference on Artificial Intelligence}, volume~40, 9368--9376.

\bibitem[{Tang et~al.(2025)Tang, Qiu, Xie, Tian, Jiao, and Ye}]{aks}
Tang, X.; Qiu, J.; Xie, L.; Tian, Y.; Jiao, J.; and Ye, Q. 2025.
\newblock Adaptive Keyframe Sampling for Long Video Understanding.
\newblock In \emph{Proceedings of the IEEE/CVF Conference on Computer Vision and Pattern Recognition}, 29118--29128.

\bibitem[{Wang et~al.(2026)Wang, Guo, Chen, Hu, and Xiong}]{qgate}
Wang, S.; Guo, W.; Chen, Z.; Hu, X.; and Xiong, H. 2026.
\newblock Where to Focus: Query-Modulated Multimodal Keyframe Selection for Long Video Understanding.
\newblock arXiv:2604.17422.

\bibitem[{Wang et~al.(2024)Wang, Zhang, Zohar, and Yeung-Levy}]{videoagent}
Wang, X.; Zhang, Y.; Zohar, O.; and Yeung-Levy, S. 2024.
\newblock Videoagent: Long-form video understanding with large language model as agent.
\newblock In \emph{European Conference on Computer Vision}, 58--76. Springer.

\bibitem[{Wang et~al.(2025)Wang, Yu, Stengel-Eskin, Yoon, Cheng, Bertasius, and Bansal}]{videotree}
Wang, Z.; Yu, S.; Stengel-Eskin, E.; Yoon, J.; Cheng, F.; Bertasius, G.; and Bansal, M. 2025.
\newblock VideoTree: Adaptive Tree-based Video Representation for LLM Reasoning on Long Videos.
\newblock In \emph{Proceedings of the IEEE/CVF Conference on Computer Vision and Pattern Recognition (CVPR)}, 3272--3283.

\bibitem[{Weng et~al.(2024)Weng, Han, He, Chang, and Zhuang}]{LongVLM}
Weng, Y.; Han, M.; He, H.; Chang, X.; and Zhuang, B. 2024.
\newblock Longvlm: Efficient long video understanding via large language models.
\newblock In \emph{European Conference on Computer Vision}, 453--470. Springer.

\bibitem[{Wu et~al.(2024)Wu, Li, Chen, and Li}]{wu2024longvideobench}
Wu, H.; Li, D.; Chen, B.; and Li, J. 2024.
\newblock Longvideobench: A benchmark for long-context interleaved video-language understanding.
\newblock \emph{Advances in Neural Information Processing Systems}, 37: 28828--28857.

\bibitem[{Yang et~al.(2026)Yang, Huang, Li, Liu, Yan, Gu, Yu, Li, and Sun}]{MemoryCard}
Yang, Q.; Huang, P.; Li, X.; Liu, Z.; Yan, Y.; Gu, Y.; Yu, G.; Li, G.; and Sun, M. 2026.
\newblock MemoryCard: Topic-Aware Multi-Modal Clue Compression for Long-Video Question Answering.
\newblock arXiv:2606.05917.

\bibitem[{Ye et~al.(2025)Ye, Wang, Sun, Chandrasegaran, Durante, Eyzaguirre, Bisk, Niebles, Adeli, Li, Wu, and Li}]{tstar}
Ye, J.; Wang, Z.; Sun, H.; Chandrasegaran, K.; Durante, Z.; Eyzaguirre, C.; Bisk, Y.; Niebles, J.~C.; Adeli, E.; Li, F.-F.; Wu, J.; and Li, M. 2025.
\newblock Re-thinking Temporal Search for Long-Form Video Understanding.
\newblock \emph{2025 IEEE/CVF Conference on Computer Vision and Pattern Recognition (CVPR)}, 8579--8591.

\bibitem[{Zhai et~al.(2023)Zhai, Mustafa, Kolesnikov, and Beyer}]{zhai2023sigmoidlosslanguageimage}
Zhai, X.; Mustafa, B.; Kolesnikov, A.; and Beyer, L. 2023.
\newblock Sigmoid Loss for Language Image Pre-Training.
\newblock arXiv:2303.15343.

\bibitem[{Zhang et~al.(2024)Zhang, Wu, Li, Li, Ma, Liu, and Li}]{zhang2410video}
Zhang, Y.; Wu, J.; Li, W.; Li, B.; Ma, Z.; Liu, Z.; and Li, C. 2024.
\newblock Video Instruction Tuning with Synthetic Data.
\newblock \emph{arXiv preprint arXiv:2410.02713}.

\end{thebibliography}


\end{document}